\documentclass[conference]{IEEEtran}
\IEEEoverridecommandlockouts
\usepackage{cite}
\usepackage{amsmath,amssymb,amsfonts}
\usepackage{algorithmic}
\usepackage{graphicx}
\usepackage{textcomp}
\usepackage{xcolor}
\def\BibTeX{{\rm B\kern-.05em{\sc i\kern-.025em b}\kern-.08em
    T\kern-.1667em\lower.7ex\hbox{E}\kern-.125emX}}
\begin{document}

\title{Skin disease identification from dermoscopy images using deep convolutional neural network}

\author{\IEEEauthorblockN{Anabik Pal$^*$}
\IEEEauthorblockA{\textit{CVPR Unit} \\
\textit{Indian Statistical Institute}\\
Kolkata-700108, India \\
E-mail: anabik\_r@isical.ac.in}
\and
\IEEEauthorblockN{Sounak Ray}
\IEEEauthorblockA{\textit{Department of Electronics and Electrical Engineering} \\
\textit{Indian Institute of Technology, Guwahati}\\
Guwahati-781039, India \\
E-mail: sounakray1997@gmail.com}
\and
\IEEEauthorblockN{Utpal Garain}
\IEEEauthorblockA{\textit{CVPR Unit} \\
\textit{Indian Statistical Institute}\\
Kolkata-700108, India \\
E-mail: utpal@isical.ac.in}
}

\maketitle

\begin{abstract}
In this paper, a deep neural network based ensemble method is experimented for automatic identification of skin disease from dermoscopic images. The developed algorithm is applied on the task3 of the ISIC 2018 challenge dataset (Skin Lesion Analysis Towards Melanoma Detection).
\end{abstract}


\section{Introduction}
The classification of pigmented skin lesions with unaided eye is challenging, even for the highly experienced dermatologists. So, dermoscopy is used for visual inspection of the skin lesions in a better way. This device can magnify the inspected regions as well as eliminates the surface reflection of the skin which leads to improve the diagnostic accuracy. But in several occasions using dermatoscope a trained experts also fail to make correct prediction~\cite{Kittler2002,Vestergaard2008}. Hence, several computer assisted automated approaches are proposed to analysis dermoscopy images~\cite{Barata2015,Alencar2016,Codella2017}. 

The identification of skin disease from dermoscopy images are treated as an image classification problem. The tradition approach of image classification needs robust feature representation which are feed to the classifier for training. So, inspired by the medical diagnostic procedure several color, texture and shape features are used to characterize the skin lesion. However, it is very much difficult to develop robust feature representation to deal with the dermoscopy images obtained from different acquisition devices and captured in diverse illumination conditions~\cite{Barata2015}. This draws the computer vision researchers to use deep convolutional neural networks.

The convolutional neural network (CNN) binds the feature extraction, feature selection and classification modules into a single unit. It can automatically extract discriminating features from the labelled images~\cite{Krizhevsky2012}. Hence, CNN produces unimaginable performance in several image classification problems. But the limitation of CNN is that it is data hungry. So, to get rid of that, transfer learning is used~\cite{Pal2016}. The transfer learning approach is nothing but tricky initialization of the network weights. In transfer learning scheme, the network weights are initialized with the leant weights of a CNN trained on another dataset. Generally, a CNN trained on the imagenet classification challenge is used for that purpose.

In this paper, ensemble of deep convolutional neural networks are used to classify the dermoscopy images into one of the seven disease classes- Melanoma, Melanocytic nevus, Basal cell carcinoma, Actinic keratosis, Benign keratosis, Dermatofibroma and Vascular lesion. We fine-tune three popular deep learning architectures namely- ResNet50~\cite{He2016}, DenseNet-121~\cite{Huang2017} and MobileNet-50~\cite{Howard2017}. to predict the disease class. Finally, a majority-voting is applied on the basis of the predicted class probability maps obtained from the trained classification networks.

\section{Dataset}

\begin{figure}[!h]
\centering
\fbox{\includegraphics[width=.18\textwidth]{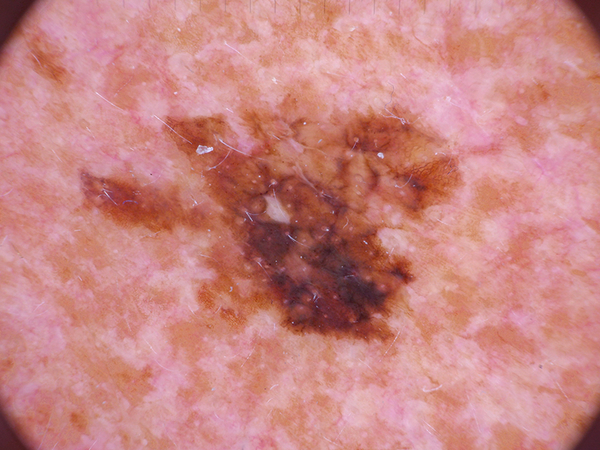}}\quad \quad \fbox{\includegraphics[width=.18\textwidth]{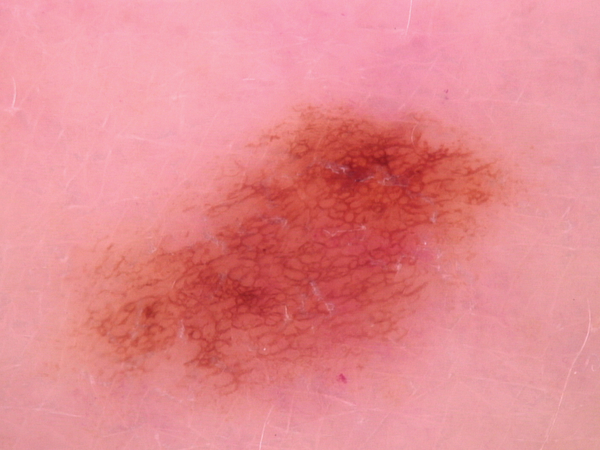}}\\
(1) Melanoma \hspace{.1\textwidth} (2) Melanocytic nevus\\
\vspace{.15cm}
\fbox{\includegraphics[width=.18\textwidth]{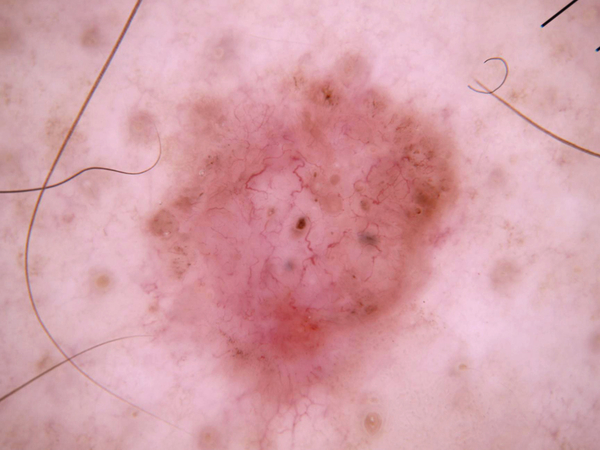}}\quad \quad \fbox{\includegraphics[width=.18\textwidth]{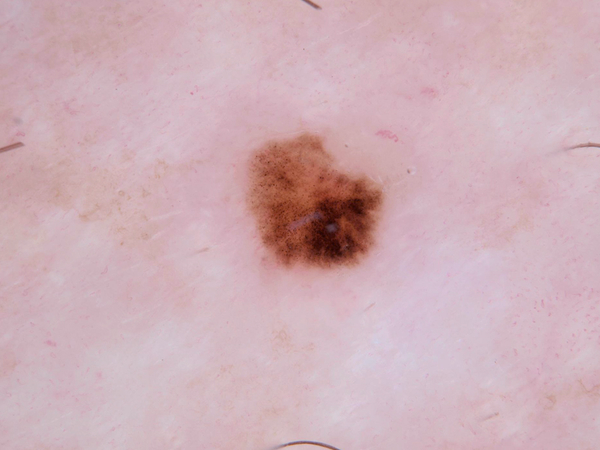}}\\
\hspace{-.02\textwidth}(3) Basal cell carcinoma  \hspace{0.05\textwidth}(4) Actinic keratosis\\
\vspace{.15cm}
\fbox{\includegraphics[width=.18\textwidth]{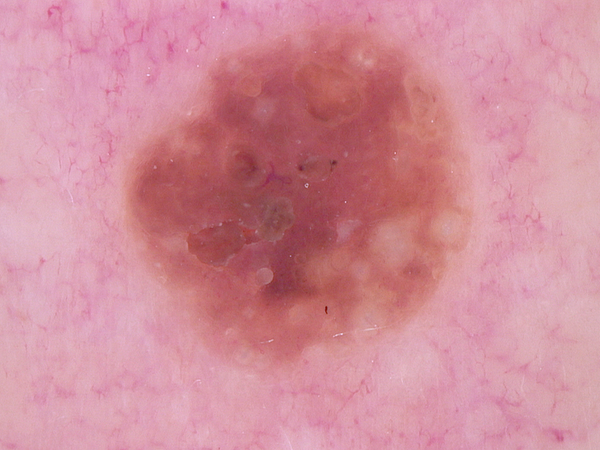}}\quad \quad \fbox{\includegraphics[width=.18\textwidth]{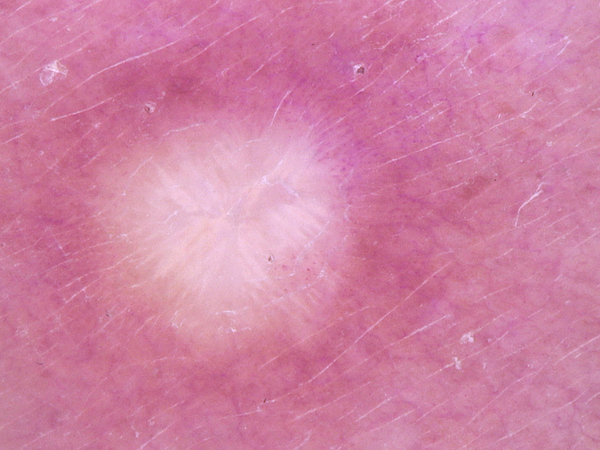}}\\
\hspace{-.075\textwidth}(5) Benign keratosis \hspace{.1\textwidth} (6) Dermatofibroma\\
\vspace{.15cm}
\fbox{\includegraphics[width=.18\textwidth]{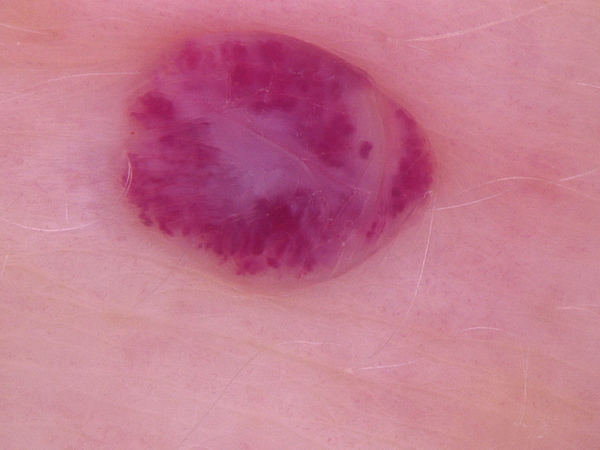}}\\
(7) Vascular\\
\caption{Images of different skin lesion.}
\label{Fig:Sample Images}
\end{figure}

In this research, the challenge dataset produced for the workshop \textbf{ISIC 2018: Skin Lesion Analysis Towards Melanoma Detection}\footnote{https://challenge2018.isic-archive.com/task3/} is used~\cite{Tschandl2018,Codella2018}. In training set, there are a total of $10015$ skin lesion images from seven skin diseases- Melanoma ($1113$), Melanocytic nevus ($6705$), Basal cell carcinoma ($514$), Actinic keratosis ($327$), Benign keratosis ($1099$), Dermatofibroma ($115$) and Vascular ($142$). The validation dataset consists of $193$ images. Sample images from all seven lesion types are shown in Figure~\ref{Fig:Sample Images}.

\section{Proposed Methodology}

In this paper, we used ensemble of three trained convolutional neural networks to identify the skin lesion. Ensemble learning is nothing but aggregation of predicted scores obtained from different classifiers. It is used for combining multiple weak classifiers to develop a stronger classifier. The individual classifiers can be constructed in several ways such as (a) using different classification algorithms, (b) training same classifier with different hyperparameters (c) using different training sets.

In this paper, we used three state-of the art convolutional neural network models namely ResNet50~\cite{He2016}, DenseNet-121~\cite{Huang2017} and MobileNet~\cite{Howard2017}. The success of these networks in image net classification challenge motivate us for choosing them. The training dataset suffers from data imbalance. We tackle this problem by back propagating the weighted loss from the loss layer. For classifier model construction, we fine-tuned the pre-trained weights of these models separately. Finally, the average predicted class probabilities obtained from these trained networks are used to decide the class label of the test image. Thus an input image is classified into one of the specified lesion classes.

\section{Implementation Details}
We used 10\% of the training images as validation images. The validation images are used for deciding the training hyper parameters. Before fine-tuning a pre-trained model, firstly the last layer (soft-max classification layer) is removed and then the number of node in last layer is set to $7$ (as we are dealing with $7$ class classification problem). Firstly, except the the last layer all other layers are freezed and the network is trained with a learning rate of 0.01, for 10 epochs with early stopping having a patience of 5 (i.e., if there is no improvement in validation loss after 5 epochs the backpropagation algorithm automatically terminates). After that, all layers are unfreezed and fine-tuned with a learning rate of 0.001 for 100 epochs. This time we used early stopping having a patience of 10. Horizontal and vertical flipping is used for augmentation.

\section{Result and Discussion}\label{Sect:Result}
The performance of the developed classifiers are scored using a normalized multi-class accuracy metric (balanced across categories). This scoring is obtained from the online portal of the challenge. The scores obtained on the validation images ($193$) are listed in Table~\ref{Table:Experimental Result}. According to Table~\ref{Table:Experimental Result}, a performance is improved when the ensembling is performed.

\begin{table}[!h]
\centering
\caption{Experimental Result}
\begin{tabular}{|c|c|c|}
\hline
Sl No.&Method&Score\\
\hline
1&ResNet-50&	0.773\\
\hline
2&DenseNet-121&	0.698\\
\hline
3&MobileNet&	0.597\\
\hline
6 & Ensemble (1,2)&  0.769\\
\hline
5&Ensemble (2,3)& 0.715\\
\hline
4&Ensemble (1,3)& 0.771\\
\hline
7&Ensemble (1,2,3)&	\textbf{0.775}\\
\hline
\end{tabular}

\label{Table:Experimental Result}
\end{table}


\section*{Acknowledgment}
We are thankful to the organizer of \textbf{\textquotedblleft MICCAI 2018 challenge ISIC} for providing the skin images.

\end{document}